%% file: arxiv.tex
\documentclass[conference,11pt]{IEEEtran}

\usepackage[utf8]{inputenc}
\usepackage[T1]{fontenc}
\usepackage{graphicx}
\usepackage{amsmath,amssymb,amsfonts}
\usepackage{booktabs}
\usepackage{hyperref}
\usepackage{subcaption}
\usepackage{xcolor}
\usepackage{url}
\usepackage{caption}
\usepackage{cite}
\usepackage{etoolbox}

\hypersetup{
  colorlinks=true,
  linkcolor=black,
  citecolor=black,
  urlcolor=black,
}

\begin{document}

\title{Prompt-Conditioned FiLM and Multi-Scale Fusion on MedSigLIP for Low-Dose CT Quality Assessment}

\author{
\IEEEauthorblockN{Tolga Demiroglu\textsuperscript{1}, Mehmet Ozan Unal\textsuperscript{1}, Metin Ertas\textsuperscript{2}, Isa Yildirim\textsuperscript{1}}
\\
\IEEEauthorblockA{\textsuperscript{1}Electronics and Communication Engineering Department, Istanbul Technical University, Istanbul, Turkey\\
\textsuperscript{2}Electrical and Electronics Engineering Department, Istanbul University, Istanbul, Turkey\\
Email: demiroglut21@itu.edu.tr, unalmehmet@itu.edu.tr, ertas@istanbul.edu.tr, iyildirim@itu.edu.tr}
}

\maketitle

\begin{abstract}
\input{abstract.tex}
\end{abstract}


\begin{figure*}[t]
  \centering
  \includegraphics[width=.99\textwidth]{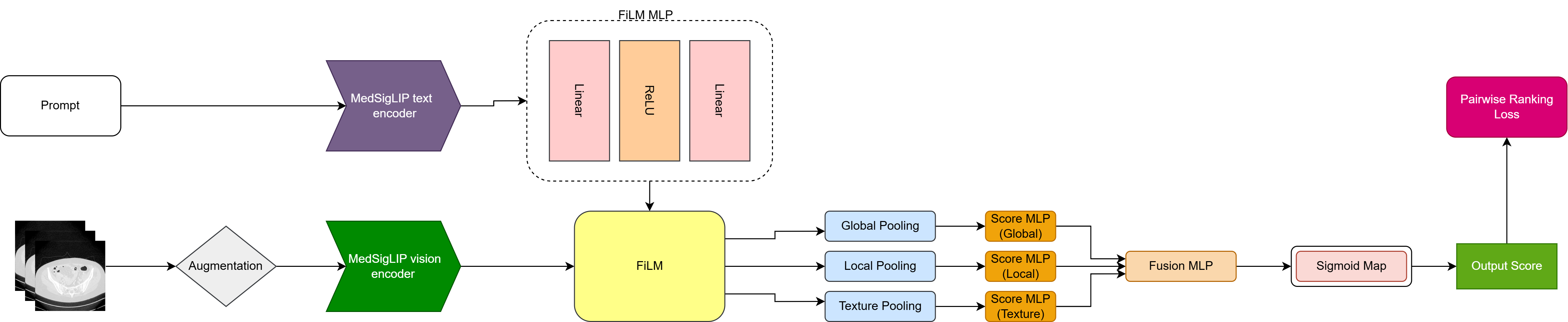}
  \caption{Prompt-conditioned FiLM with multi-scale (global/local/texture) pooling.}
  \label{fig:architecture}
\end{figure*}

\section{Introduction}
Low-dose computed tomography (LDCT) is clinically necessary to reduce radiation exposure; however, lowering dose increases quantum noise, streak artifacts, and texture washout, degrading perceived image quality and diagnostic confidence. Paired LDCT-NDCT datasets are scarce due to ethical constraints, and reference-based metrics show fundamental limitations in medical imaging~\cite{breger2025study}, making \emph{reference-free image quality assessment} essential for LDCT.

Existing reference-free approaches face key challenges: \textit{(i)} limited labeled Mean Opinion Scores (MOS) require generalizable learning from few annotations, \textit{(ii)} successful methods need large datasets and long training, and \textit{(iii)} natural-image metrics fail to capture clinical context where acceptable noise varies by anatomy.

We address these gaps with a \emph{prompt-conditioned} quality assessment framework built on \emph{MedSigLIP}~\cite{sellergren2025medgemma}. We inject clinical intent via text prompts through Feature-wise Linear Modulation (FiLM)~\cite{perez2018film}, where MedSigLIP's frozen text encoder transforms the prompt into scale/shift parameters $(\gamma,\beta)$ that modulate patch-token features: $\tilde{\mathbf{h}} = \mathbf{h} \odot (1 + \alpha \cdot \tanh(\gamma)) + \alpha \cdot \beta$. The conditioned embeddings are aggregated via three pooling strategies (global, local 4-region, 2-bin max for texture), feeding separate regression heads fused by an MLP. Predictions are mapped to $[0,4]$ via temperature-scaled sigmoid and trained with pairwise ranking loss.

On LDCTIQA2023 challenge~\cite{lee2025low} (1,300 images: 1,000 train, 300 test; radiologist-assigned MOS 0--4), our method surpasses the best published model with \textbf{PLCC = 0.9575}, \textbf{SROCC = 0.9561}, \textbf{KROCC =0.8301}. Our contributions in this study can be detailed as follows: (\textit{i}) a MedSigLIP-based, prompt-guided formulation; (\textit{ii}) FiLM-based text injection with multi-scale pooling; (\textit{iii}) a pairwise ranking loss tailored to MOS ordering.
\section{Related Work}

\subsection{Language-Guided LDCT Methods}
Recent work explores language priors in LDCT restoration. LEDA uses LLM supervision for LDCT$\rightarrow$NDCT denoising~\cite{chen2024low}, A-IDE employs LLM-based expert routing to anatomy-specialized denoisers~\cite{cho2025ide}, and LangMamba integrates VLM representations into autoencoder-based denoising~\cite{chen2025langmamba}. IQAGPT applies LLM-prompted criteria for CT quality assessment~\cite{chen2024iqagpt}. However, these methods optimize reconstruction fidelity or operate post-hoc, lacking token-level conditioning inside the vision backbone.

\subsection{Positioning of Our Work}
In contrast to denoising/reconstruction pipelines above, we target
\emph{reference-free quality scoring} for LDCT. We employ the frozen pre-trained MedSigLIP vision encoder as our feature extractor, then apply FiLM (conditioned on text prompts) to the encoded patch tokens output, and aggregate via multi-scale (global-local-texture) pooling followed by a lightweight fusion head trained predominantly with a ranking loss.
This yields a bounded $[0,4]$ score aligned with clinical rubrics, adapts quickly by editing
prompts (rather than re-training reconstruction modules), and can serve as a
plug-in criterion for tuning or auditing these restoration systems.

\section{Method}

\subsection{Problem Setup}
Given a low-dose CT slice $I \in \mathbb{R}^{H \times W}$ and a textual instruction (prompt) $t$, our goal is to produce a bounded quality score $\hat y \in [0,4]$ reflecting perceived diagnostic quality under the intent specified by $t$. We use MedSigLIP to encode images and text, with an image backbone $f_{\mathrm{img}}$ and a text encoder $f_{\mathrm{text}}$:
\begin{equation}
z_I = f_{\mathrm{img}}(I), \qquad z_t = f_{\mathrm{text}}(t).
\end{equation}
Unless otherwise noted, the text tower is frozen to improve data efficiency.

\subsection{Architecture Overview}

We adopt the MedSigLIP vision backbone and inject prompt information via FiLM applied to the \emph{final} patch-token features. Let $H \in \mathbb{R}^{B \times P \times d}$ denote the last hidden states over $P$ patch tokens (batch size $B$, embedding dimension $d{=}1152$). After applying FiLM, the conditioned tokens are aggregated through three parallel pooling branches: (i) a \emph{global} branch (average pool) for overall quality trends, (ii) a \emph{local} branch (4-region average pool) to preserve spatial heterogeneity, and (iii) a \emph{texture} branch (2-bin max pool) emphasizing worst-case artifacts. Each pooled representation feeds a dedicated regression head, producing sub-scores $y_{\mathrm{g}}, y_{\mathrm{l}}, y_{\mathrm{tex}}\in\mathbb{R}$ that are fused by a two-layer MLP to yield the final logit, mapped to $[0,4]$ via temperature-scaled sigmoid (see Figure~\ref{fig:architecture}).

\subsection{Prompt-Conditioned FiLM}
We condition the token features using FiLM parameters \((\gamma,\beta)\) predicted from the prompt embedding~\cite{perez2018film}. 
Let $z_t\!\in\!\mathbb{R}^{d_t}$ be the normalized text embedding and let $d$ denote the channel width of the vision tokens.
A two-layer MLP $g(\cdot):\mathbb{R}^{d_t}\!\to\!\mathbb{R}^{2d}$ maps $z_t$ to these channel-wise scale/shift parameters:
\begin{equation}
(\gamma,\beta)=g(z_t), \qquad \gamma,\beta\in\mathbb{R}^{d}.
\end{equation}
Let $s$ denote a scalar FiLM strength. Following our implementation, FiLM is applied with a bounded scale via $\tanh(\cdot)$:
\begin{equation}
\widetilde{H} = H \odot \bigl(1 + s \cdot \tanh(\gamma)\bigr) + s \cdot \beta,
\end{equation}
where the affine transformation is broadcast across tokens. The modulated features $\widetilde{H}$ are used by all subsequent heads.

\subsection{Multi-Scale Global-Local-Texture Pooling and Fusion}
Let $\tilde H\!\in\!\mathbb{R}^{B\times P\times d}$ be the FiLM-modulated patch tokens and
$V=\tilde H^\top\!\in\!\mathbb{R}^{B\times d\times P}$.
We extract three complementary summaries corresponding to \emph{global}, \emph{local}, and \emph{texture} branches:
\begin{equation}
\begin{aligned}
h_{\mathrm{g}}   &= \mathrm{AvgPool}_{1}(V)  \in \mathbb{R}^{B\times d},\\
h_{\mathrm{l}}   &= \mathrm{AvgPool}_{4}(V)  \in \mathbb{R}^{B\times 4d},\\
h_{\mathrm{tex}} &= \mathrm{MaxPool}_{2}(V)  \in \mathbb{R}^{B\times 2d}.
\end{aligned}
\end{equation}
Branch-specific heads $\psi_{\mathrm{g}},\psi_{\mathrm{l}},\psi_{\mathrm{tex}}$ map these summaries to sub-scores
$y_{\mathrm{g}},y_{\mathrm{l}},y_{\mathrm{tex}}\in\mathbb{R}$. The final prediction is obtained by fusing the sub-scores:
\begin{equation}
\text{logit}=\phi\!\bigl([y_{\mathrm{g}}\;\|\;y_{\mathrm{l}}\;\|\;y_{\mathrm{tex}}]\bigr),\qquad
\hat y = 4\,\sigma\!\bigl(\text{logit}/\tau_{\mathrm{out}}\bigr).
\end{equation}
Here, \(\mathrm{AvgPool}_{1}\) summarizes global context (overall noise), 
\(\mathrm{AvgPool}_{4}\) preserves regional details (edges, streaks), and 
\(\mathrm{MaxPool}_{2}\) emphasizes worst-case texture regions.

\subsection{Pairwise Ranking Loss}
We optimize a pairwise ranking loss with an optional regression term. 
Let $\mathcal{P}=\{(i,j)\,:\, y_i\neq y_j\}$ be the set of ordered, non–tied pairs in a mini-batch of predictions $\{\hat y_i\}$ and targets $\{y_i\}$, and define $s_{ij}=\mathrm{sign}(y_i-y_j)\in\{-1,+1\}$. 
Following RankNet \cite{burges2005learning}, the pairwise logistic loss with temperature $\tau_{\mathrm{rank}}{=}0.5$ is
\begin{equation}
\mathcal{L}_{\mathrm{rank}}
=\frac{1}{|\mathcal{P}|}\!\sum_{(i,j)\in\mathcal{P}}
\log\!\left(1+e^{\frac{-s_{ij}\,(\hat y_i-\hat y_j)}{\tau_{\mathrm{rank}}}}\right),
\end{equation}
(i.e., $\mathrm{softplus}(-\,s_{ij}(\hat y_i-\hat y_j)/\tau_{\mathrm{rank}})$ averaged over pairs). In implementation, we mask out ties ($y_i{=}y_j$) and average the remaining pairwise terms, which only differs by a constant scale.
The optional regression term is the mean–squared error
\begin{equation}
\mathcal{L}_{\mathrm{mse}}=\tfrac{1}{B}\sum_{i=1}^{B}(\hat y_i-y_i)^2.
\end{equation}
The total loss combines both terms:
\begin{equation}
\mathcal{L}_{\mathrm{total}}=\lambda_{\mathrm{rank}}\,\mathcal{L}_{\mathrm{rank}}
+\lambda_{\mathrm{mse}}\,\mathcal{L}_{\mathrm{mse}}.
\end{equation}
In our main runs, we set $\lambda_{\mathrm{rank}}{=}1$ and 
$\lambda_{\mathrm{mse}}{=}0$, effectively using pure pairwise ranking. This 
choice is empirically motivated: preliminary experiments showed that pairwise 
loss alone outperformed pure MSE by enforcing relative quality ordering, which 
is more robust under MOS annotation noise. Mixed weighting (e.g., small 
$\lambda_{\mathrm{mse}}{>}0$) may offer further gains by combining absolute 
and relative supervision, but is deferred to future work.

\section{Experiments}\label{sec:experiments}
\subsection{Dataset}

We evaluate our method on the LDCTIQA2023 challenge~\cite{lee2025low}, which provides 1,000 training images and 300 test images with radiologist-assigned MOS scores on a 0--4 rubric
(\emph{0: bad; 1: poor; 2: fair; 3: good; 4: excellent}).
We train on the provided training set and report results on the official \emph{test} set using the challenge metrics: Pearson (PLCC), Spearman (SROCC), and Kendall (KROCC).

\subsection{Implementation Details}
We use the \texttt{google/medsiglip-448} checkpoint as backbone; images are resized to $448{\times}448$. 
The default prompt consists of key quality attributes:
\begin{quote}\small\ttfamily
Rate this low-dose CT (MOS 0-4): 0 Nondiagnostic—desired features not shown; 1 Poor—diagnostic interpretation impossible; 2 Fair—limited \linebreak interpretation; 3 Good—diagnostic; 4 Excellent—anatomy highly visible. Return only one number 0-4.
\end{quote}
 For optimization, we use AdamW with learning rate $1{\times}10^{-5}$, weight decay $1{\times}10^{-4}$, and cosine annealing scheduler, along with batch size $4$ and gradient
accumulation $\times 2$. Mixed precision (AMP) is enabled. The text tower is frozen; FiLM strength
is $s{=}1.0$. Predictions are mapped to $[0,4]$ via $\hat y = 4\,\sigma(\text{logit}/\tau_{\text{out}})$; we set
$\tau_{\text{out}}{=}2.0$ to soften the sigmoid and reduce edge compression bias. Data augmentation includes random horizontal flip
($p{=}0.5$), small rotation ($\pm 10^\circ$), and mild brightness/contrast jitter. 

We perform 5-fold cross-validation on the 1,000 training images; each fold is trained for 22 epochs with checkpoints saved based on validation loss. For final model selection, we use validation MAE rather than ranking loss, as ranking loss can overfit to relative orderings while MAE directly measures absolute score prediction accuracy. The fold with lowest validation MAE was selected as the final model. Code is available at \url{https://github.com/itu-biai/medsiglip_ldct_iqa}.

\subsection{Main Results on LDCTIQA2023 Test Set}

We report results on the official test set (300 images) using the challenge's evaluation metrics~\cite{lee2025low}: Pearson (PLCC), Spearman (SROCC), and Kendall (KROCC), along with their sum (Overall).
For evaluation, all test images are resized to $448{\times}448$ without augmentation.

\newcommand{\best}[1]{\textbf{#1}\,{\small$\uparrow$}}

\begin{table}[h]
\centering
\caption{\centering Quantitative comparison on the LDCTIQA2023 \textbf{test} set. Best in \textbf{bold} with $\uparrow$.}
\label{tab:ldctiqa_test}
\small
\setlength{\tabcolsep}{2.5pt}
\begin{tabular}{p{0.35\columnwidth}cccc}
\toprule
Methods & PLCC  & SROCC  & KROCC  & Overall  \\
\midrule
\textbf{Ours} & \best{0.9575} & \best{0.9561} & 0.8301 & \best{2.7436} \\
1st           & 0.9491 & 0.9495 & 0.8440 & 2.7427 \\
2nd           & 0.9434 & 0.9414 & 0.7995 & 2.6843 \\
3rd           & 0.9402 & 0.9387 & 0.7930 & 2.6719 \\
4th           & 0.9362 & 0.9338 & 0.7851 & 2.6550 \\
5th           & 0.9278 & 0.9232 & 0.7691 & 2.6202 \\
\bottomrule
\end{tabular}
\end{table}

\noindent
Table~\ref{tab:ldctiqa_test} compares our method with the top-ranked submissions reported by the challenge~\cite{lee2025low}.
Our approach attains the highest PLCC, SROCC, and Overall, with slightly lower KROCC than the 1st method.
These results highlight the effectiveness of MedSigLIP's vision backbone for medical image quality assessment, suggesting potential for broader applications with task-specific prompt engineering.

\subsection{Ablation Studies}\label{sec:ablations}

\noindent\textbf{Effect of prompt relevance and FiLM.}
We analyze sensitivity to the text prompt and FiLM. Using an \emph{intentionally irrelevant} prompt, we compare:
(1) FiLM \emph{on} ($s{=}1$), (2) FiLM \emph{off} ($s{=}0$) with the same prompt (thus no text injection), and
(3) our final model with a clinically aligned prompt and FiLM on.

\begin{table}[htbp]
  \caption{\centering Ablation on prompt relevance and FiLM (LDCTIQA2023 test).}
  \centering
  \small
  \setlength{\tabcolsep}{1.2pt}
  \begin{tabular}{p{0.41\columnwidth}cccc}
    \toprule
    \textbf{Setting} & \textbf{PLCC} & \textbf{SROCC} & \textbf{KROCC} & \textbf{Overall} \\
    \midrule
    FiLM on + irrelevant prompt & 0.9487 & 0.9485 & 0.8137 & 2.7109 \\
    FiLM off (no prompt) & 0.9517 & 0.9507 & 0.8167 & 2.7192 \\
    FiLM on + clinical prompt (Ours) & \textbf{0.9575} & \textbf{0.9561} & \textbf{0.8301} & \textbf{2.7436} \\
    \bottomrule
  \end{tabular}
  \label{tab:ablation_prompt_film}
\end{table}

\noindent\textbf{Irrelevant prompt used.}
For the "irrelevant prompt" condition we intentionally used a non-medical, aesthetic description unrelated to CT quality:
\begin{quote}\small\ttfamily
Describe the visual beauty of blooming flowers in a spring garden, focusing on colors, petals, and the gentle \linebreak sunlight. Use poetic language to evoke emotion.
\end{quote}

\noindent
Table~\ref{tab:ablation_prompt_film} shows that poorly chosen prompts may not provide gains when injected via FiLM.
When prompt quality is uncertain, reducing $s$ or disabling FiLM ($s{=}0$) is recommended.
Conversely, a clinically aligned prompt combined with FiLM ($s{=}1$) yields consistent gains.

\noindent\textbf{Effect of the loss function (MSE vs. Pairwise Ranking Loss).}
We compare a pure MSE loss against our pairwise ranking loss.

\begin{table}[htbp]
  \caption{\centering Ablation on the loss function (LDCTIQA2023 test).}
  \label{tab:ablation_loss}
  \centering
  \small
  \setlength{\tabcolsep}{2.5pt}
  \begin{tabular}{p{0.34\columnwidth}cccc}
    \toprule
    \textbf{Loss} & \textbf{PLCC} & \textbf{SROCC} & \textbf{KROCC} & \textbf{Overall} \\
    \midrule
    MSE only  & 0.9411 & 0.9425 & 0.8044 & 2.6880 \\
    \textbf{Pairwise Ranking} & \best{0.9575} & \best{0.9561} & \best{0.8301} & \best{2.7436} \\
    \bottomrule
  \end{tabular}
\end{table}

\noindent
Table~\ref{tab:ablation_loss} shows that pairwise ranking loss significantly outperforms MSE (2.7436 vs. 2.6880 Overall), confirming its effectiveness in preserving relative quality ordering under limited annotations.

\section{Conclusion}
We presented a prompt-conditioned quality assessment model for LDCT built on MedSigLIP.
Text cues are injected via FiLM applied to the vision encoder's output patch embeddings,
then combined with global-local-texture token pooling and a lightweight fusion head to predict a bounded score.

On the LDCTIQA2023 benchmark, our method achieved
\textbf{PLCC}=0.9575, \textbf{SROCC}=0.9561,
\textbf{KROCC}=0.8301, and \textbf{Overall}=2.7436,
with consistent gains in ablations (FiLM off / prompt variants).
The approach is data-efficient and adapts quickly to new intents through prompting.

Limitations include MOS subjectivity, prompt sensitivity, and dataset/scanner diversity.
Future work will investigate pooling architecture variants, multi-layer FiLM, instruction-style prompts, uncertainty calibration,
and multi-center validation.

\apptocmd{\thebibliography}{\setlength{\itemsep}{0pt}}{}{}
\bibliographystyle{IEEEtran}
\bibliography{refs}

\end{document}

%% file: abstract.tex
 We propose a prompt-conditioned framework built on MedSigLIP that injects textual priors via Feature-wise Linear Modulation (FiLM) and multi-scale pooling. Text prompts condition patch-token features on clinical intent, enabling data-efficient learning and rapid adaptation. The architecture combines global, local, and texture-aware pooling through separate regression heads fused by a lightweight MLP, trained with pairwise ranking loss. Evaluated on the LDCTIQA2023 (a public LDCT quality assessment challenge) with 1,000 training images, we achieve PLCC = 0.9575, SROCC = 0.9561, and KROCC = 0.8301, surpassing the top-ranked published challenge submissions and demonstrating the effectiveness of our prompt-guided approach~\cite{lee2025low}.